%% file: neurips_2026.tex
\documentclass{article}

\PassOptionsToPackage{numbers, compress}{natbib}
 \usepackage[preprint]{neurips_2026}

\usepackage[utf8]{inputenc} 
\usepackage[T1]{fontenc}    
\usepackage{hyperref}       
\usepackage{url}            
\usepackage{booktabs}       
\usepackage{amsfonts}       
\usepackage{nicefrac}       
\usepackage{microtype}      
\usepackage{xcolor}         
\usepackage{enumitem}
\usepackage{amsmath}
\usepackage{graphicx}
\usepackage{tabularx}  
\usepackage{array}
\usepackage{mdframed}

\usepackage{multirow}  

\title{PhysAgent: Automating Physics-Based 4D Synthesis via Trajectory-Grounded Multi-Agent Feedback}

\author{
  Chunji Lv, Jiaxi Ye, Yuchen Jiang, Rexar Lin, Changsheng Li\thanks{Corresponding author} \\
  Beijing Institute of Technology \\
  \texttt{3120250994@bit.edu.cn} 
}

\begin{document}

\maketitle

\input{section/abstract}
\input{section/introduction}
\input{section/relatedwork}
\input{section/method}

\input{section/experiments}
\input{section/conclusion}

\clearpage
\bibliographystyle{unsrtnat}
\bibliography{myrefs}


\newpage
\appendix 
\renewcommand{\thefigure}{A\arabic{figure}}
\setcounter{figure}{0}

\input{section/appendix}

\end{document}

%% file: section/abstract.tex
\begin{abstract}
Achieving fully automated, physically plausible 3D motion synthesis is a core objective in graphics and generative AI. However, configuring complex environmental force fields still relies entirely on manual expert intervention, creating a severe bottleneck for large-scale simulation data generation. Existing automated methods primarily focus on material optimization and exhibit severe modality gaps and technical flaws when applied to the vastly more complex force field optimization space: naive Large Language Models (LLMs) lack underlying simulation feedback, causing severe physical inaccuracies, while traditional Score Distillation Sampling (SDS) suffers from sluggish gradients, local optima entrapment, and a mathematical inability to dynamically switch discrete force fields. To address this, we propose PhysAgent, the first simulator-in-the-loop multi-agent framework that leverages multimodal inputs for automated, physically grounded 4D synthesis. By decoupling intrinsic materials from extrinsic dynamics, PhysAgent utilizes a Semantic Agent equipped with an externalized Force Field Skill module to master simulation rules and generate valid initializations. Subsequently, the Refine Agents—driven by Trajectory-Grounded Multi-Agent Feedback—leverage vision foundation models to extract dense point trajectories from rendered frames. By converting these explicit motion trajectories into structured textual descriptors, the agent harnesses LLM commonsense reasoning to execute zero-shot macroscopic leaps, effectively escaping local optima and dynamically switching discrete force fields. Extensive experiments demonstrate that PhysAgent rapidly generates stable, diverse physical scenes from arbitrary multimodal prompts, significantly outperforming existing baselines in both generation diversity and physical accuracy.
\end{abstract}

%% file: section/introduction.tex
\section{Introduction}

\begin{figure}[th]
\centering
\vspace{-5pt}
\includegraphics[width=0.9\linewidth]{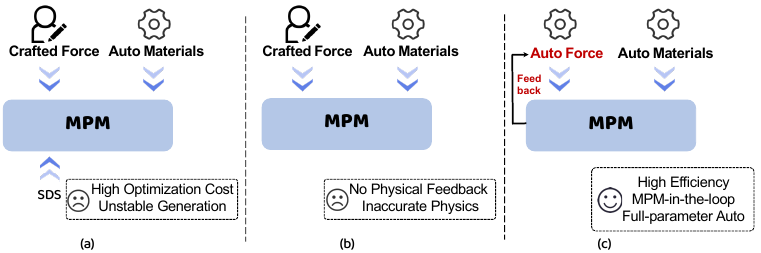}

\caption{\textbf{Comparison of physics-based 4D synthesis paradigms.} (a) SDS-based methods automate material optimization but suffer from high optimization cost, unstable generation and rely on manually crafted forces. (b) Naive generative approaches automate materials but lack physical feedback, leading to inaccuracies and neglecting environmental forces. (c) Our PhysAgent proposes a "simulator-in-the-loop" paradigm. By automating both force fields and materials with continuous visual feedback, it achieves full-parameter automation, high efficiency, and physical compliance.}
\label{fig:head}
\vspace{-12pt}
\end{figure}

The pursuit of fully automated, physically grounded 4D synthesis represents a holy grail in computer graphics and generative AI. While recent physics-based 4D synthesis methods have demonstrated remarkable progress, they predominantly focus on optimizing the geometric and material properties of individual objects~\cite{xie2024physgaussian,lv2025physgm,huang2025dreamphysics,lin2025omniphysgs,zhang2024physdreamer}. However, in realistic physical simulations that empower embodied AI~\cite{black2024pi_0,kim2024openvla}, the dynamic evolution of diverse scenes is fundamentally governed by environmental force fields. Currently, these force field parameters heavily rely on manual configuration by experts, creating a severe bottleneck for large-scale simulation data generation. Therefore, we argue that the fully automated synthesis of complex and dynamic environment force fields represents a highly critical yet largely unexplored frontier for achieving truly automated and diverse 4D physics synthesis.

Unfortunately, bridging the modality gap between high-level user intent and low-level physical parameters remains a formidable challenge. Current optimization strategies generally exhibit critical bottlenecks. On the one hand, traditional optimization pipelines~\cite{huang2025dreamphysics,lin2025omniphysgs} relying on differentiable simulator-based Score Distillation Sampling (SDS)~\cite{poole2022dreamfusion} are heavily constrained. SDS relies on sluggish gradient backpropagation, is inherently prone to being trapped in local optima, and is mathematically incapable of handling the dynamic switching of discrete variables (e.g., toggling between different types of force fields). On the other hand, naive Large Language Model (LLM) prompting attempts to directly map image to physical parameters~\cite{zhao2025physsplat}. However, without simulation feedback, LLMs fail to grasp the complex physical dimensions of MPM and the underlying simulation mechanisms, inevitably leading to severe physical distortions. Therefore, we argue that existing methods (Figure~\ref{fig:head} (a,b)) are insufficient to reliably translate user intent into dynamic parameters, necessitating a novel framework that simultaneously transcends the mathematical rigidity of backpropagation and the physical hallucinations of ungrounded LLMs.

To shatter these rigid bottlenecks, we propose PhysAgent , a multi-agent framework that accepts joint text-and-image multimodal inputs to synthesize 4D physical dynamics. Rather than confounding geometry and dynamics, our framework extracts initial intrinsic materials and spatial anchors from the reference image, while automatically configuring the complex environmental force fields strictly from the user's text prompt. As shown in Figure~\ref{fig:head} (c), by thoroughly abandoning the slow gradients of traditional SDS and the physical inaccuracies of naive LLM prompting, we design a novel simulator-in-the-loop multi-agent generative paradigm. Specifically, we first design a Semantic Agent capable of interpreting user inputs. Crucially, we equip this agent with an externalized Force Field Skill module, which explicitly defines the taxonomy and operational constraints of all supported force types. This not only ensures the robustness and physical compliance of the initial planning, but also fundamentally resolves the pain point of traditional LLMs' inability to comprehend Material Point Method(MPM)~\cite{jiang2016material,stomakhin2013material} numerical characteristics and underlying simulation mechanisms. Subsequently, we propose the core of our framework: the Refine Agents, driven by a Trajectory-Grounded Multi-Agent Feedback mechanism. To construct a reliable visual perception loop, this mechanism leverages vision foundation models (e.g., SAM 2~\cite{kirillov2023segment,ravi2024sam}, DepthAnythingv3~\cite{yang2024depth,depthanything3} and CoTracker3~\cite{karaev2025cotracker3,karaev2024cotracker}) to extract precise object motion trajectories from rendered frames. These explicit kinematic trajectories provide fine-grained visual dynamic priors that are perfectly orthogonal to the LLMs' semantic reasoning. Driven by these precise visual feedback signals, our multi-agent mechanism operates as a closed loop : a Planning agent analyzes trajectory deviations to issue optimization instructions, while an Executor agent iteratively updates the force field configurations until convergence--unlike SDS-based methods, the Refine Agent leverages global commonsense reasoning to execute zero-shot macroscopic state leaps, effortlessly escaping local optima. Furthermore, it can dynamically switch discrete force field types, thereby entirely bypassing the sluggish gradient backpropagation and suboptimal updates inherent to traditional optimization.

In summary, our main contributions are as follows:
\begin{itemize}[leftmargin=*]

\item We propose \textbf{PhysAgent}—the first MPM-simulator-in-the-loop multi-agent framework for automated and diverse 4D physical synthesis. By explicitly decoupling image-derived intrinsic priors and text-driven extrinsic dynamics, it pioneers the rapid generation of complex environmental force field configurations purely from semantic intents.

\item We design a Semantic Agent equipped with an externalized ``Force Field Skill module'', enabling a physically plausible initialization of the physical force fields.

\item We propose Refine Agents based on trajectory-grounded feedback. By utilizing vision foundation models and harnessing the global reasoning capabilities of LLMs, this agent effectively bypasses the sluggish optimization of SDS, enabling zero-shot macroscopic leaps to escape local optima.

\item Extensive experiments demonstrate that PhysAgent significantly outperforms existing baseline methods in both generation diversity and physical accuracy.
\end{itemize}

%% file: section/relatedwork.tex
\section{Related Work}
\subsection{Generative 4D Gaussian Splatting}
Driven by advancements in 3D Gaussian Splatting (3DGS)~\cite{kerbl20233d}, 4D dynamic scene generation primarily relies on data-driven approaches. These typically train native generative models on 3D representations~\cite{jun2023shap,nichol2022point,nash2017shape,chen2023single,tang2024lgm,hong2023lrm,ziwen2025long,zhang2024gs,zhu2024motiongs} or extract priors from 2D diffusion models via score distillation~\cite{ling2024align,poole2022dreamfusion,ren2023dreamgaussian4d,yin2024improved,yin20234dgen,zhang20244diffusion,jiang2024animate3d,pan2026efficient4d}. Despite incorporating simple kinematic trajectories~\cite{wang2025physctrl,gao2026lome} or heuristics~\cite{gillman2025force}, these fundamentally data-driven architectures lack explicit physical modeling. Consequently, they often produce "physical hallucinations"—visually coherent dynamics that violate real-world physical laws. In contrast, our framework operates strictly upon rigorous physical simulation to ensure superior physical fidelity.

\subsection{Physics-Grounded 4D Gaussian Splatting}
To inject realistic dynamics, recent milestones have coupled physical simulators like the Material Point Method (MPM)~\cite{macklin2022warp,jiang2016material,stomakhin2013material} with 3DGS. While PhysGaussian~\cite{xie2024physgaussian,cao2026physgaussian} pioneered this, its manual parameter tuning restricts scalability. Subsequent works used Score Distillation Sampling (SDS) to automate object material learning~\cite{huang2025dreamphysics,lin2025omniphysgs,tan2024physmotion,liu2024physics3d,zhang2024physdreamer,wang2026motionphysics}, but suffer from excruciating optimization times and remain confined to intrinsic properties. Methods leveraging LLMs~\cite{zhao2025physsplat,chen2025physgen3d,liu2024physgen}, feed-forward Gaussians~\cite{lv2025physgm}, or multi-material configurations~\cite{liu2026mosiv,ma2026fastphysgs,zhang2025physchoreo} accelerate this process but collectively overlook environmental force fields. Existing physics-based paradigms face two critical limitations: (1) an object-centric focus that neglects indispensable dynamic environmental force fields, and (2) a reliance on computationally prohibitive gradient backpropagation through differentiable simulators or SDS, which is highly susceptible to local optima.

\subsection{LLM Agents for Physical Reasoning}

The rise of Large Language Models (LLMs) has inspired the development of autonomous agents for code generation~\cite{wang2023voyager,singh2022progprompt}, tool utilization~\cite{schick2023toolformer}, and physical task planning~\cite{ma2023eureka}. While recent works have explored LLM agents for robotic control~\cite{cui2025grove}, or high-level 3D layout generation~\cite{feng2023layoutgpt,yang2024holodeck}, their application to complex continuum mechanics (e.g., MPM) is notably absent due to the severe modality disconnect between discrete semantics and continuous physical states. Naive open-loop prompting for simulation~\cite{zhao2025physsplat,chen2025physgen3d,xue2025phyt2v} is fundamentally flawed; without visual grounding, these "blind" agents cannot comprehend trajectory deviations. PhysAgent bridges this gap via a novel simulator-in-the-loop paradigm. By formalizing a Force Field Skill module and leveraging vision models for multi-agent feedback, we translate kinematic priors into actionable reasoning, bypassing both sluggish gradient optimization and physically distorted open-loop generation.

%% file: section/method.tex
\section{Method}
\label{sec:method}

\begin{figure}[t]
\centering
\vspace{-5pt}
\includegraphics[width=\linewidth]{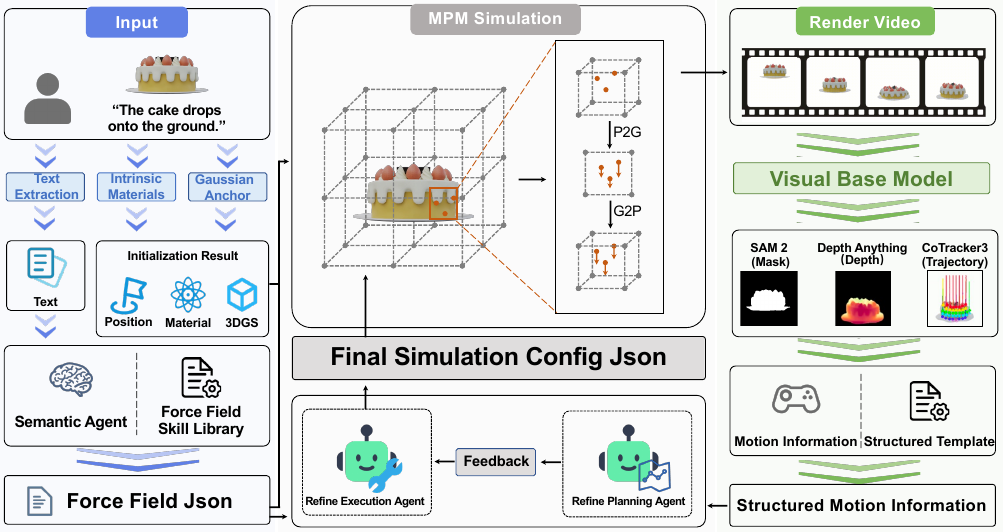}

\caption{\textbf{PhysAgent Framework Overview.}  Operating in a closed-loop ``simulator-in-the-loop" paradigm, the system processes multimodal inputs. First, intrinsic materials and Gaussian anchors are extracted from the reference image to initialize the 3DGS representation and object physical properties. Concurrently, the Semantic Agent interprets the text prompt and queries the Force Field Skill Library to generate the force field json. These initialization results are deterministically fused into a final simulation config json for MPM simulation. Visual foundation models then extract segment, depth, and trajectory modalities from the rendered video to form structured motion feedback. Finally, the Refine Agents analyze this feedback to optimizing the configuration.}
\label{fig:pipeline}
\vspace{-12pt}
\end{figure}

This section details the PhysAgent framework. As illustrated in Figure \ref{fig:pipeline}, PhysAgent is designed to take joint multimodal inputs (an image and a corresponding text prompt) and output physically grounded 4D motion videos. Operating under a novel ``simulator-in-the-loop'' multi-agent paradigm, our framework translates the user intent into dynamic environmental force field configurations for 4D synthesis. We first formalize our core optimization problem (Sec. \ref{sec:problem_formulation}). Subsequently, we detail the robust initialization via the Semantic Agent (Sec. \ref{sec:semantic}), followed by the closed-loop Trajectory-Grounded Multi-agent Optimization (Sec. \ref{sec:optimization}) that unifies physical simulation, visual perception, and reasoning-driven refinement.

\subsection{Problem Formulation}
\label{sec:problem_formulation}
The primary objective of PhysAgent is to translate user intent into dynamic environmental force field configurations, thereby driving fully automated and physically grounded 4D motion synthesis.

Formally, we denote a multimodal prompt as $\mathcal{P}_{multi} = \{\mathcal{P}_{image}, \mathcal{P}_{text}\}$, where $\mathcal{P}_{image}$ grounds the initial geometry and material priors, and $\mathcal{P}_{text}$ directs the dynamic evolution. The underlying vector force field configuration is denoted as $\Theta$. To ensure that our optimization space strictly converges on the environmental force field design rather than being confounded by material parameters, we decouple the intrinsic physical properties. Specifically, the intrinsic material properties of the 3D assets are pre-acquired from $\mathcal{P}_{image}$ utilizing PhysGM~\cite{lv2025physgm}, denoted as $\Phi_{\text{PhysGM}}$.

Furthermore, to bridge the explicit 3D Gaussian Splatting (3DGS) representation with the continuous MPM simulation grid, we implement a lightweight spatial anchoring module. This extracts the initial spatial positions (i.e., the mean vectors $\mu$) of the 3D Gaussians, denoted as $\mathbf{s}_0$, which explicitly define the valid action regions for the applied force fields.

Unlike traditional differentiable simulations relying on continuous gradient-based loss minimization, our framework operates in a semantic, gradient-free optimization paradigm. Let $\text{Sim}(\cdot)$ represent the MPM simulator and $\mathcal{E}_{visual}(\cdot)$ denote the visual perception module that extracts structured kinematic behaviors. The core optimization problem is formulated as finding the optimal force field configuration $\Theta^*$ maximizing the semantic and physical alignment with the user's text intent $\mathcal{P}_{text}$:
\begin{equation}
\Theta^* = \arg\max_{\Theta} \mathcal{A}_{\text{LLM}}\Big(\mathcal{E}_{visual}\big(\text{Sim}(\mathbf{s}_0, \Theta, \Phi_{\text{PhysGM}})\big), \mathcal{P}_{text}\Big)
\end{equation}
where $\mathcal{A}_{\text{LLM}}(\cdot, \cdot)$ is a alignment evaluation function inherently modeled by our Refine Agents. This formulation converts the optimization into a closed-loop causal reasoning process, dynamically updating $\Theta$ to satisfy the target physical constraints without computing scalar numerical losses.

\subsection{Text-Driven Semantic Agent}
\label{sec:semantic}
Translating high-level natural language into low-level simulation parameters poses a significant challenge, as naive Large Language Models (LLMs) lack an inherent understanding of continuous physical states and numerical stability boundaries. To address this, we design a Text-Driven Semantic Agent, strictly guided by an externalized force field skill library. 

Taking only the user's text instruction $\mathcal{P}_{text}$ as input, the Semantic Agent queries this library, which explicitly defines the taxonomy, mathematical formulations, and operational constraints of all supported macroscopic forces. Acting as a semantic compiler, the agent generates a structured force field json, encapsulating the discrete force types and their corresponding continuous parameters. The detailed design of the force field skill library is presented in the appendix. 

However, a standalone force field is insufficient to launch a comprehensive MPM simulation. To construct the complete initial simulation environment, the generated force field must be deterministically fused with explicit physical and spatial priors. Specifically, we systematically integrate four distinct components through the initialization module: (1) the semantic force field json, (2) the 3DGS representation, (3) the intrinsic materials $\Phi_{\text{PhysGM}}$, and (4) the Gaussian spatial coordinate anchors $\mathbf{s}_0$. This fusion mechanism generates the initial grounded simulation configuration JSON ($\Theta_0$). It rigorously ensures that the initial state is physically compliant, spatially accurate.

\subsection{Forward Physical Simulation via Material Point Method}
\label{sec:mpm_simulation}
Once initialized, the framework enters a closed-loop refinement phase. To physically synthesize realistic 4D dynamics from the current configuration $\Theta_k$, we employ the Material Point Method (MPM). Functioning as the forward execution engine of our loop, MPM is a hybrid Lagrangian-Eulerian approach that handles deformations robustly. 

The continuum body is discretized into Lagrangian \textit{particles} carrying physical properties. During the particle-to-grid (\textbf{P2G}) transfer, mass $m$ and momentum are mapped to grid nodes $i$ using interpolation shape functions $N_i(\mathbf{x}_p)$:
\begin{equation}
m_i = \sum_p m_p N_i(\mathbf{x}_p), \quad (m\mathbf{v})_i = \sum_p m_p \mathbf{v}_p N_i(\mathbf{x}_p)
\end{equation}
Next, the momentum is updated on the grid by solving Newton's second law:
\begin{equation}
\mathbf{v}_i^{t+\Delta t} = \mathbf{v}_i^t + \Delta t m_i^{-1} \left( \mathbf{f}_i^{int} + \mathbf{f}_i^{ext}(\Theta_k) \right)
\end{equation}
where $\mathbf{f}_i^{int}$ denotes internal elastic forces. Crucially, $\mathbf{f}_i^{ext}(\Theta_k)$ represents the external environmental forces injected by our framework. It is precisely at this grid-solving phase that our explicitly defined macroscopic fields act to alter the kinematic states. Finally, in the grid-to-particle (\textbf{G2P}) step, the updated grid velocities and positions are interpolated back to advance the particles:
\begin{equation}
\mathbf{v}_p^{t+\Delta t} = \sum_i \mathbf{v}_i^{t+\Delta t} N_i(\mathbf{x}_p), \quad \mathbf{x}_p^{t+\Delta t} = \mathbf{x}_p^t + \Delta t \mathbf{v}_p^{t+\Delta t}
\end{equation}
The simulator subsequently drives the 3DGS to render a 2D physical video sequence $V$.

\subsection{Trajectory-Grounded Multi-agent Feedback and Optimization}
\label{sec:optimization}
Building upon the forward physical simulation, the framework integrates visual kinematic perception and reasoning-driven parameter updates to iteratively optimize the force field configuration $\Theta_k$ towards the user's intent.

\paragraph{Trajectory-Grounded Visual Perception}
While the underlying MPM simulator explicitly maintains precise 3D states for millions of micro-particles, directly exposing these dense coordinates to the LLM agent would lead to severe information overload. Furthermore, relying on the simulator's privileged 3D ground truth inherently limits the framework to closed white-box environments. 

To bridge this sim-to-real gap, we intentionally treat the simulator as a black-box 2D renderer. Given the rendered video $V = \{I_1, I_2, \dots, I_T\}$, we employ SAM 2 to generate frame-wise binary masks $\mathbf{M}_t$. To overcome the depth ambiguity of 2D tracking, we leverage Depth Anything V3 to extract dense monocular depth maps $\mathbf{D}_t$. Combined with camera intrinsics $\mathbf{K}$, the masked 2D features are deterministically lifted to 3D coordinates:
\begin{equation}
\mathbf{p}_t^{(i)} = \mathbf{K}^{-1} [x_t^{(i)}, y_t^{(i)}, 1]^\top \cdot \mathbf{D}_t(x_t^{(i)}, y_t^{(i)})
\end{equation}
where $(x_t^{(i)}, y_t^{(i)}) \in \mathbf{M}_t$. CoTracker3 then temporally tracks these unprojected features, yielding consistent 3D trajectories $\mathcal{T}$. A formatting module synthesizes these macroscopic trajectories into Structured Motion Information ($\mathcal{F}_k$), acting as the extracted "visual loss" for downstream agents.

\paragraph{Reasoning-Driven Configuration Refinement}
Acting as the "optimizer" of the loop, the Refine Agents process the feedback to execute parameter updates. This module is decoupled into the Refine Planning Agent and the Refine Execution Agent. 

At the $k$-th iteration, the Planning Agent evaluates the alignment between the extracted motion $\mathcal{F}_k$ and the original text intent $\mathcal{P}_{text}$ via physical commonsense reasoning. If the kinematics match the target dynamics, the simulation terminates. Otherwise, it formulates targeted text-based feedback. 

Guided by this feedback, the Execution Agent updates the configuration. Unlike traditional differentiable methods (e.g., Score Distillation Sampling) that rely on sluggish local updates $\Theta_{k+1} = \Theta_k - \eta \nabla \mathcal{L}_{\text{SDS}}$, our agent evaluates the causal relationship subject to physical constraints $\mathcal{C}_{physics}$:
\begin{equation}
\Theta_{k+1} = \mathcal{G}_{\text{LLM}}(\Theta_k, \mathcal{F}_k, \mathcal{P}_{text}, \mathcal{C}_{physics})
\end{equation}
Guided by the semantic mapping function $\mathcal{G}$, the Execution Agent performs zero-shot macroscopic leaps. This closed-loop process bypasses the local optima inherent to numerical optimization.

%% file: section/experiments.tex
\section{Experiments}
\vspace{-5pt}
\subsection{Implementation Details}
We implement PhysAgent using the PyTorch framework. Both the 3D Gaussian Splatting (3DGS) representations and the intrinsic material properties are initialized utilizing PhysGM. To capture fine-grained physical dynamics, the subsequent Material Point Method (MPM) simulations are conducted on a high-resolution $100^3$ background grid. During the initialization phase, we employ a density-based particle filling strategy, capping the maximum number of simulated particles at 200,000 to achieve the balance between highly realistic physical deformation and computational efficiency.

Regarding the temporal simulation design, we synthesize 50 continuous frames for each dynamic sequence, setting the frame interval ($\Delta t_{\text{frame}}$) to $4 \times 10^{-2}$ seconds. Crucially, to ensure numerical stability within the continuous physical space and prevent numerical explosions during extreme force interactions, the MPM simulator operates at an extremely fine-grained sub-step ($\Delta t_{\text{sub}}$) resolution of $2 \times 10^{-4}$ seconds. The base simulation environment is initialized with a default scaled gravity and a standard frictionless collision plane; these base configurations are subsequently dynamically overridden or augmented by our multi-agent framework.

We select Qwen3.5 9B~\cite{qwen3.5} as the backbone for both the Semantic Agent and the reasoning-driven Refine Agents. The visual perception module employs the pre-trained weights of SAM 2, Depth Anything V3, and CoTracker3. All experiments---including baseline comparisons and our physical simulations---are strictly executed on a single NVIDIA RTX PRO 5000 GPU to ensure fairness.

\begin{figure}[t]
\centering
\vspace{-5pt}
\includegraphics[width=0.9\linewidth]{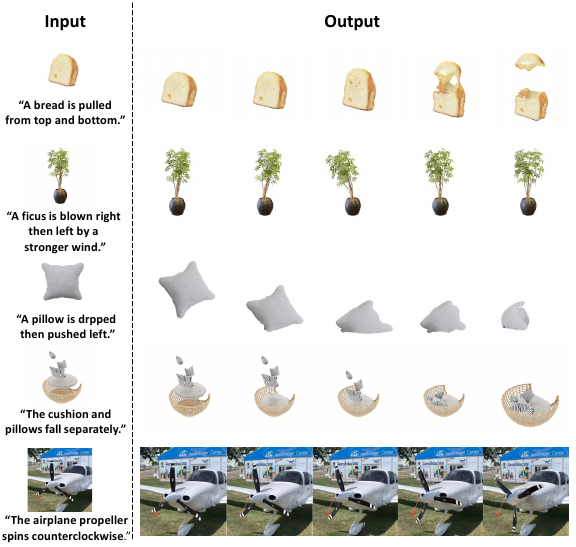}
\caption{Qualitative results. Continuous physical dynamic responses generated by PhysAgent under various text prompts.}
\label{fig:exp1}
\vspace{-12pt}
\end{figure}

\vspace{-5pt}
\subsection{Datasets}
To evaluate the performance of PhysAgent, we conduct experiments on two distinct datasets: 

\textbf{PhysGaussian~\cite{xie2024physgaussian}:} Following recent works, we utilize the static scenes from the PhysGaussian dataset to ensure a fair baseline comparison. For these scenes, the inputs to our framework consist of the provided 3D Gaussian Splatting (3DGS) representations paired with custom text prompts.

\textbf{Objaverse~\cite{deitke2023objaverse}:} To evaluate the generalization capability of our framework across diverse, large-scale 3D assets, we curate a representative subset of 3D models from the Objaverse dataset for dynamic demonstration. For this dataset, our framework accepts joint multimodal inputs, specifically utilizing reference images alongside custom text prompts.

\subsection{Comparison with State-of-the-Art Methods}
\label{subsec:compare}
\textbf{Baselines.} We compare PhysAgent against recent state-of-the-art physics-based 4D synthesis methods: PhysSplat, DreamPhysics, and OmniPhysgs. Because these baselines predominantly focus on intrinsic material optimization, we adopt two distinct evaluation strategies to ensure a rigorously fair comparison: (1) \textit{Expert-Crafted vs. Auto-Generated}: We align all other physical parameters and compare our autonomously generated force fields against the manually crafted force field configurations provided by their domain experts. (2) \textit{Adapted Optimization vs. Multi-Agent Reasoning}: We adapt their material optimization techniques (e.g., SDS) to the optimization space of environmental force fields, effectively evaluating our reasoning-driven configurations against those produced by existing gradient-based optimization methods.

\textbf{Metrics.} We employ two primary quantitative metrics. To assess the semantic alignment between the user's text prompt and the generated dynamic video, we utilize \textbf{CLIPsim}~\cite{radford2021learning}. To evaluate physical realism and visual naturalness, we conduct a user study and report the \textbf{User Preference Rating (UPR)}. Additionally, we provide comprehensive qualitative visualizations for direct comparison.

\begin{figure}[b]
\centering
\vspace{-8pt}
\includegraphics[width=0.9\linewidth]{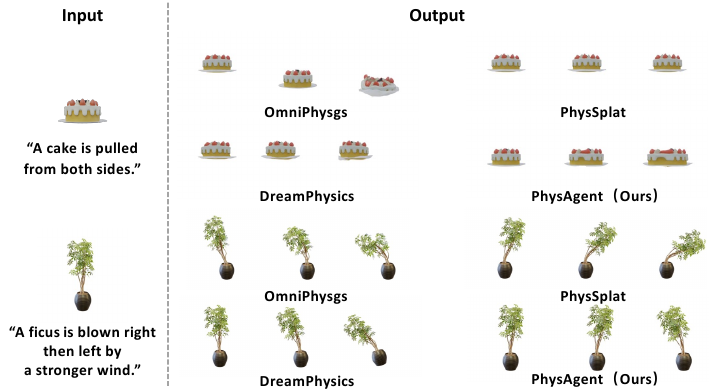}

\caption{Qualitative comparisons. Comparison of dynamic responses generated by PhysAgent and baseline methods given complex force descriptions (e.g., pulling and alternating wind).}
\label{fig:exp2}

\end{figure}

\textbf{Qualitative Results.} As illustrated in Figure \ref{fig:exp1}, PhysAgent successfully synthesizes highly realistic dynamic simulation sequences driven by diverse multimodal inputs. In comparison with baseline models (Figure \ref{fig:exp2}), existing methods exhibit significant limitations when processing complex multimodal instructions. Specifically, baselines relying on Score Distillation Sampling (SDS) or open-loop generation frequently suffer from severe physical distortions under complex environmental force fields, manifesting as erroneous force applications or implausible physical phenomena. This fundamentally stems from their lack of explicit physical constraints on force field parameters and their inherent difficulty in optimizing force directions that possess discrete decision attributes. In stark contrast, PhysAgent is capable of generating exceptionally smooth and physically plausible deformations.

\begin{table}[!t]

\centering
\caption{Quantitative comparisons. We evaluate our method and baseline models on 3 different force types. Evaluation is based on the $\text{CLIP}_{sim}$ score (higher is better $\uparrow$) and UPR (higher is better $\uparrow$).}
\small

\renewcommand{\arraystretch}{1.3}

\begin{tabularx}{\textwidth}{l *{8}{>{\centering\arraybackslash}X}}
\toprule 
\multirow{2}{*}{\raisebox{-0.5\normalbaselineskip}{Method}} & \multicolumn{2}{c}{drop} & \multicolumn{2}{c}{stretching} & \multicolumn{2}{c}{sway} & \multicolumn{2}{c}{average} \\
\cmidrule(lr){2-3} \cmidrule(lr){4-5} \cmidrule(lr){6-7} \cmidrule(lr){8-9}
& $\text{CLIP}_{sim}$ & UPR & $\text{CLIP}_{sim}$ & UPR & $\text{CLIP}_{sim}$ & UPR & $\text{CLIP}_{sim}$ & UPR \\
\midrule 

OmniPhysgs-f~\cite{lin2025omniphysgs} 
& 0.2612 & 15\% & 0.2507 & 0\% & 0.2751 & 10\% & 0.2623 & 8.3\% \\
OmniPhysgs-g~\cite{lin2025omniphysgs} 
& 0.2625 & 40\% & 0.2605 & 40\% & 0.2772 & 40\% & 0.2667 & 40\% \\
OmniPhysgs~\cite{lin2025omniphysgs}+PhysAgent
& \textbf{0.2627} & \textbf{45\%} & \textbf{0.2715} & \textbf{60\%} & \textbf{0.2805} & \textbf{50\%} & \textbf{0.2716} & \textbf{51.7\%} \\
\midrule
PhysSplat-f~\cite{zhao2025physsplat}
& 0.2998 & 25\% & 0.2968 & 10\% & 0.2547 & 5\% & 0.2838 & 13.3\% \\
PhysSplat-g~\cite{zhao2025physsplat}
& 0.3057 & 35\% & 0.3029 & 40\% & 0.2897 & 35\% & 0.2994 & 36.7\% \\
PhysSplat~\cite{zhao2025physsplat}+PhysAgent
& \textbf{0.3070} & \textbf{40\%} & \textbf{0.3163} & \textbf{50\%} & \textbf{0.2911} & \textbf{60\%} & \textbf{0.3048} & \textbf{50\%} \\

\midrule

DreamPhysics-f~\cite{huang2025dreamphysics}
& 0.2712 & 5\% & 0.2751 & 10\% & 0.2574 & 10\% & 0.2679 & 8.3\% \\
DreamPhysics-g~\cite{huang2025dreamphysics}
& 0.2873 & 25\% & 0.2959 & 35\% & 0.2937 & 40\% & 0.2923 & 33.3\% \\
DreamPhysics~\cite{huang2025dreamphysics}+PhysAgent
& \textbf{0.3090} & \textbf{70\%} & \textbf{0.3041} & \textbf{55\%} & \textbf{0.3264} & \textbf{50\%} & \textbf{0.3132} & \textbf{58.3\%} \\
\bottomrule 
\end{tabularx}

\label{tab:table1}
\vspace{-8pt}
\end{table}

\textbf{Quantitative Results.} Table \ref{tab:table1} validates PhysAgent's superiority across three evaluation settings: ``-f'' (adapting baseline optimization to the force field space), ``-g'' (expert-annotated baseline force configurations), and ``+PhysAgent'' (integrating our force field into baselines). PhysAgent consistently achieves the highest $\text{CLIP}_{sim}$ and User Preference Rating (UPR) scores. Notably, it exhibits substantial gains in complex interactive tasks like ``Stretching'' and ``Sway,'' proving that trajectory-grounded feedback effectively mitigates physical hallucinations. 

The lower performance of baselines stems from three critical bottlenecks: (1) Expert annotations (``-g'') frequently fall into sub-optimal states due to human subjectivity and the lack of iterative visual feedback. (2) Traditional SDS optimization is highly susceptible to local optima when handling discrete force attributes (e.g., toggling directions). (3) Naive open-loop LLMs suffer from severe physical hallucinations, failing to comprehend MPM numerical dimensions without explicit simulator grounding. Conversely, PhysAgent leverages LLM global reasoning and rigorous simulator feedback to overcome these limitations, ensuring physically plausible and near-perfect semantic alignment.

\subsection{Ablation Study}

\begin{figure}[h]
\centering

\includegraphics[width=\linewidth]{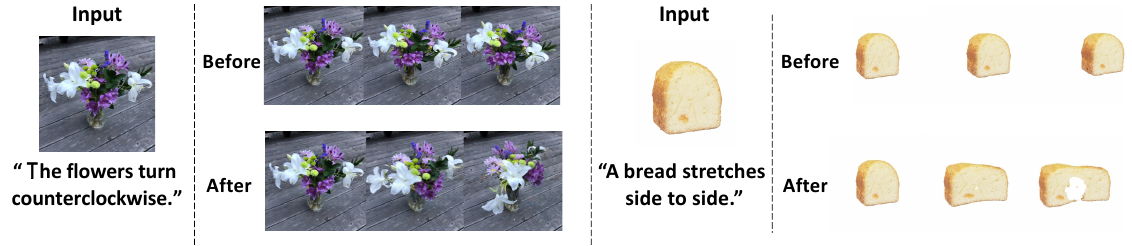}

\caption{Qualitative ablation of the Refine Agents. ``Before'' denotes the open-loop generation, which often leads to insufficient deformation or suboptimal dynamics. ``After'' demonstrates that introducing our closed-loop Refine Agents successfully rectifies these kinematic deviations.}
\label{fig:ablation}

\end{figure}

To thoroughly investigate the contributions of each core component within PhysAgent, we conduct comprehensive ablation studies. The quantitative and qualitative results are presented in Table \ref{tab:ablation} and Figure \ref{fig:ablation}, respectively.

\noindent\textbf{Necessity of the Force Field Skill Library.} As indicated in Table \ref{tab:ablation}, removing the externalized Force Field Skill module (\textit{w/o Skill}) leads to a precipitous degradation in overall system performance. This demonstrates that, despite the formidable commonsense reasoning capabilities of LLMs, the absence of explicit physical rule constraints and well-defined parameter spaces inevitably induces severe physical parameter hallucinations. Consequently, the model becomes incapable of correctly configuring the low-level parameters strictly required by the MPM simulator.

\noindent\textbf{Critical Role of the Refine Agents.} Eliminating the optimization agent (\textit{w/o Refine Agent}) degrades the framework into a naive open-loop generation paradigm. Figure \ref{fig:ablation} visually illustrates this contrast: relying solely on the Semantic Agent's initial predictions (Before) frequently results in anomalous kinematic drifting and suboptimal dynamics. Conversely, incorporating the Refine Agent (After) empowers the system to perform iterative optimizations driven by explicit visual feedback (i.e., trajectory deviations) extracted from the simulator. Quantitatively, as shown in Table \ref{tab:ablation}, the integration of the Refine Agent yields substantial improvements across all evaluation metrics. This compellingly substantiates that ``trajectory-grounded feedback'' is the pivotal pathway for achieving high-fidelity physical synthesis, as it enables the model to continuously rectify the deviations of its initial semantic predictions within the continuous physical space.

\begin{table}[t] 
\centering
\vspace{-5pt}
\caption{Quantitative ablation study. Comparison across different force types, demonstrating the necessity of the Force Field Skill Library and the Refine Agents for high-fidelity 4D synthesis.}
\label{tab:ablation}
\renewcommand{\arraystretch}{1.2}

\resizebox{\linewidth}{!}{
\begin{tabular}{l *{8}{c}}
\toprule 
\multirow{2}{*}{\raisebox{-0.5\normalbaselineskip}{Method}} & \multicolumn{2}{c}{drop} & \multicolumn{2}{c}{stretching} & \multicolumn{2}{c}{sway} & \multicolumn{2}{c}{average} \\
\cmidrule(lr){2-3} \cmidrule(lr){4-5} \cmidrule(lr){6-7} \cmidrule(lr){8-9}
& $\text{CLIP}_{sim}$ $\uparrow$ & UPR $\uparrow$ & $\text{CLIP}_{sim}$ $\uparrow$ & UPR $\uparrow$ & $\text{CLIP}_{sim}$ $\uparrow$ & UPR $\uparrow$ & $\text{CLIP}_{sim}$ $\uparrow$ & UPR $\uparrow$ \\
\midrule 
w/o Force Field Skill &0.2601 & 10\%& 0.2869& 5\%& 0.2560& 15\%& 0.2677&10\% \\
w/o Refine Agents      &0.2919 & 25\%& 0.3271& 25\%& 0.3092& 30\%& 0.3094&26.7\% \\
\midrule
\textbf{PhysAgent (Full)} & \textbf{0.3023} & \textbf{65\%} & \textbf{0.3348} & \textbf{70\%} & \textbf{0.3110} & \textbf{55\%} & \textbf{0.3160} & \textbf{63.3\%} \\
\bottomrule 
\end{tabular}
}
\vspace{-8pt}
\end{table}

%% file: section/conclusion.tex
\vspace{-10pt}
\section{Discussion}
PhysAgent fills a critical gap in 4D physics synthesis by automating extrinsic environmental force fields. Because both force configurations and traditional material properties are optimized as structured parameters within the MPM, our framework ensures theoretical rigor and inherent compatibility with existing material-centric methods. Currently, to balance simulation fidelity with computational cost, our MPM grid is limited to a $100^3$ resolution. Future research will focus on distilling this ``simulator-in-the-loop" framework into a unified, native "Physics-Vision-Language" foundation model, which would bypass intermediate perception modules to enable real-time, large-scale dynamic generation in open environments.  
\vspace{-10pt}
\section{Conclusion}
We propose PhysAgent, a fully automated multi-agent framework that utilizes multimodal inputs for physically grounded 4D motion synthesis. By explicitly decoupling image-derived intrinsic material priors from text-driven environmental force fields, it bridges the gap between discrete semantics and continuous physical states. Operating within a novel simulator-in-the-loop paradigm, PhysAgent leverages a Force Field Skill library and visual trajectory feedback to empower LLMs to execute zero-shot macroscopic state leaps. Extensive experiments demonstrate that the framework effortlessly escapes local optima, eliminates physical hallucinations, and achieves robust, human-level 4D physical simulations.

%% file: section/appendix.tex
\noindent \textbf{\Large Appendix}

\section{Semantic Agent and Force Field Skill Library}
\label{sec:force_field_skill}

To bridge the modality gap between natural language and MPM numerical constraints, PhysAgent designs a Qwen3.5-powered Semantic Agent. Acting as a ``Vector Field Parser'' it translates physical intents into standardized JSONs for compiling continuous mathematical force fields.

\subsection{Generalized Prompt for Semantic Agent}
\label{subsec:generalized_prompt}

To effectively guide the LLM's physical reasoning, we designed a structured system prompt, whose core logic is outlined below. In practice, a robust JSON extraction pipeline enables the LLM to utilize Chain-of-Thought (CoT) reasoning to decompose complex interactions before outputting the configuration.

\begin{mdframed}[backgroundcolor=gray!10, roundcorner=5pt, linewidth=0pt, innertopmargin=10pt, innerbottommargin=10pt]
\small
\textbf{Role:} Physics-Grounded Vector Field Compiler\\
\textbf{Task:} Translate natural language descriptions of physical behaviors into continuous mathematical vector parameters and temporal constraints compatible with the Material Point Method (MPM) simulator.

\textbf{1. Output Schema Definition}\\
Your final configuration must be formatted as a JSON block matching the following schema. Multiple actions can be overlaid or sequenced within the \texttt{actions} array.
\begin{verbatim}
{
    "default_drop": boolean,
    "actions": [
        {
            "action_type": "translation" | "scale" | "impulse" | "torque",
            "vector": [x, y, z],
            "magnitude": float,
            "active_time": [start_time, end_time]
        }
    ]
}
\end{verbatim}

\textbf{2. Physical Field Mapping \& MPM Integration}
\begin{itemize}[leftmargin=*]
    \item \textbf{translation}: For pushing, pulling, or blowing. Mapped to a continuous external force field ($\mathbf{f}_{ext}$) accumulated over the grid update phase.
    \item \textbf{scale}: For squeezing or stretching. Mapped to a spatially-varying force field scaling along the normal axis. \texttt{magnitude} $> 0$ denotes outward stretching; $< 0$ denotes inward compression.
    \item \textbf{impulse}: For sudden strikes, impacts, or explosions. Mapped to an instantaneous velocity modification ($\Delta \mathbf{v}$) applied directly to the MPM particles. Typically occupies a minimal temporal window (e.g., \texttt{[0, 1]}).
    \item \textbf{torque}: For twisting or stirring. Mapped to a tangential force field inducing pure rotational momentum around the specified center axis.
\end{itemize}

\textbf{3. Parameter Constraints}
\begin{itemize}[leftmargin=*]
    \item \textbf{Coordinate System:} The \texttt{vector} must be strictly resolved and normalized into a 3D unit vector.
    \item \textbf{Magnitude ($M$):} The auxiliary value of $M$ ranges from \(0.4\) to \(1.6\). Default is 1.0. Modulate monotonically based on linguistic intensity modifiers (e.g., ``slightly'' $\to$ lower bound, ``violently'' $\to$ upper bound). Reverse sign for opposing semantics (Push/Stretch $\to M > 0$, Pull/Squeeze $\to M < 0$).
    \item \textbf{Temporal Window:} Define \texttt{active\_time} based on semantic sequence. Default continuous actions run for the full sequence (e.g., \texttt{[0, 50]}).
\end{itemize}

\textbf{4. Execution Constraint}\\
You may reason step-by-step to analyze coordinates and intensities. However, the final mechanical parameters MUST be enclosed within a valid JSON block.
\end{mdframed}

\subsection{Force Field Skill Library Formulations}
\label{subsec:skill_library_math}

The \textbf{Force Field Skill Library} serves as the deterministic compiler that bridges the semantic JSON outputs and the continuous MPM simulator. It defines a standardized set of executable ``skills,'' compiling the parsed \texttt{action\_type}, normalized \texttt{vector} ($\hat{\mathbf{v}}$), and \texttt{magnitude} ($M$) into either external force fields $\mathbf{f}_i^{ext}(\Theta_k)$ or velocity modifiers. Let $\mathbf{x}_c$ denote the geometric center of the target object. 

Crucially, to bridge the dimensionless outputs of the LLM with the rigorous physical dimensions of the simulator, each skill internally defines a specific dimensional scaling factor $k_{type}$. This factor directly translates the parsed semantic magnitude $M$ into usable physical units, ensuring seamless execution within the continuous MPM space. 

\paragraph{Skill 1: Translation.}
Represents a continuous directional force (e.g., wind, pushing). The force applied to a particle at position $\mathbf{x}_p$ is uniform across the specified object mask:
\begin{equation}
\mathbf{f}_{ext}(\mathbf{x}_p) = k_{trans} \cdot M \cdot \hat{\mathbf{v}}
\end{equation}

\paragraph{Skill 2: Scale (\texttt{scale}).}
Simulates squeezing or stretching along a specific normal axis $\hat{\mathbf{v}}$. To ensure alignment with the discrete constraints generated by the Refine Agent (which treats scaling as two symmetrically opposed translations), this skill is formulated as a piecewise uniform force field divided by the object's center plane. Let $d = (\mathbf{x}_p - \mathbf{x}_c) \cdot \hat{\mathbf{v}}$ be the signed projection distance. The external force is defined using the sign function $\text{sgn}(d)$:
\begin{equation}
\mathbf{f}_{ext}(\mathbf{x}_p) = k_{scale} \cdot M \cdot \text{sgn}(d) \cdot \hat{\mathbf{v}}
\end{equation}
When $M < 0$ (squeezing), particles on opposite sides of the center plane are pushed uniformly towards it. When $M > 0$ (stretching), they are pulled uniformly outwards, effectively executing a paired translation.

\paragraph{Skill 3: Impulse.}
Represents a catastrophic, instantaneous energy injection (e.g., an explosion or hit). Unlike continuous skills, the impulse modifies the particle velocity directly over a strictly bounded, minimal time window $\Delta t_{impulse}$. Utilizing our MPM \texttt{Impulse\_modifier}, the instantaneous velocity update for particle $p$ with mass $m_p$ is:
\begin{equation}
\mathbf{v}_p^{t+\Delta t} = \mathbf{v}_p^t + \frac{k_{impulse} \cdot M \cdot \hat{\mathbf{v}}}{m_p} \Delta t
\end{equation}

\paragraph{Skill 4: Torque.}
Applies rotational dynamics around the specified axis $\hat{\mathbf{v}}$ passing through $\mathbf{x}_c$. We first compute the perpendicular radial vector $\mathbf{r}_{\perp} = (\mathbf{x}_p - \mathbf{x}_c) - \left((\mathbf{x}_p - \mathbf{x}_c) \cdot \hat{\mathbf{v}}\right)\hat{\mathbf{v}}$. The tangential force direction is defined by the cross product, yielding the external force:
\begin{equation}
\mathbf{f}_{ext}(\mathbf{x}_p) = k_{torque} \cdot M \cdot \left(\hat{\mathbf{v}} \times \mathbf{r}_{\perp}\right)
\end{equation}
This mathematically guarantees that the skill induces a pure rotational momentum around the semantic axis.

\subsection{Deterministic Configuration Fusion for MPM Initialization}
\label{subsec:config_fusion}

While the Semantic Agent exclusively handles the extrinsic dynamics (force fields), a complete MPM simulation requires a unified environment containing geometric, material, and global boundary definitions. To achieve this, our framework employs a deterministic fusion module that aggregates the decoupled modalities into a single, simulation-ready physical configuration JSON. 

The fusion process integrates four primary components:
\begin{enumerate}[leftmargin=*]
    \item \textbf{Geometric Anchoring (3DGS to Particles):} The explicit 3D Gaussian Splatting (3DGS) representation is converted into Lagrangian particles. The Gaussian means ($\boldsymbol{\mu}$) are directly mapped to initial particle positions ($\mathbf{x}_p$). The particle volumes ($V_p$) and densities are derived from the Gaussian scales ($\mathbf{s}$) and opacities ($\alpha$). These initialized arrays are cached and loaded into the MPM state.
    \item \textbf{Intrinsic Materials:} The material properties extracted via PhysGM are mapped to the constitutive models defined in our simulator. This includes the material type index (e.g., 0 for jelly, 1 for metal, 2 for sand) and its corresponding Young's modulus ($E$) and Poisson's ratio ($\nu$). 
    \item \textbf{Extrinsic Force Fields:} The compiled skills (translation, scale, impulse, torque) and their temporal constraints (\texttt{active\_time}) generated by the Semantic Agent are directly injected.
    \item \textbf{Global Hyperparameters \& Boundaries:} Default simulation attributes are appended to ensure numerical stability. This includes the background grid resolution (\texttt{n\_grid} $= 100$), grid limits, sub-step size ($\Delta t_{sub}$), base gravitational acceleration, and default boundary conditions (e.g., a frictional ground collision plane).
\end{enumerate}

\section{Refine Agents Workflow}
\label{sec:refine_agent_details}

Unlike traditional feed-forward generation pipelines, PhysAgent incorporates a ``simulation-in-the-loop'' paradigm. The Refine Agent functions as the core reasoning engine within this loop, continuously evaluating and adjusting the physical parameters until the synthesized 4D dynamics align with the user's textual prompt. 

The Refine Agent operates through a collaborative two-stage workflow: Visual Perception and Configuration Optimization.

\subsection{Visual Perception Module}
The Visual Perception module bridges the gap between raw pixel streams and high-level semantic reasoning. It is powered by a robust vision foundation pipeline comprising SAM 2, Depth Anything V3, and CoTracker3. 

\paragraph{Workflow.} 
Given the rendered RGB video stream $V_t$ from the current iteration $t$, SAM 2 provides robust binary segmentation masks of the target object. These masks strictly bound the sampling space for CoTracker3, filtering out background noise and ensuring that trajectory tracking is heavily focused on the target's intrinsic motion. Concurrently, Depth Anything V3 lifts these dense 2D tracking points into 3D representations. 

\paragraph{Semantic Feedback Generation.}
By analyzing the temporal 3D trajectories, the module extracts structured metrics, including centroid displacement, 3D rotation, and bounding box volume scaling. These metrics are summarized into a concise \textbf{Text Feedback Prompt}. For instance, a typical feedback snippet reads: \textit{``At frame 49, the centroid translates significantly with minor rotation. The bounding box volume shrinks to $0.35\times$ of the initial frame.''}. This acts as the observational evidence for the subsequent optimization phase.

\subsection{Configuration Optimization Module}
Utilizing Qwen3.5-9B as the cognitive backbone, the Configuration Optimization module processes the textual feedback to execute causal reasoning and update the MPM boundary conditions ($\mathcal{F}_t$). It operates in two sequential phases:

\paragraph{Phase 1: Evaluation and Judgment.}
The LLM acts as a strict physics reviewer. It compares the initial User Prompt against the Observation (Text Feedback) and the Context (current configuration JSON $\mathcal{F}_t$). It outputs a strict binary judgment (\texttt{SUCCESS} or \texttt{FAIL}). If \texttt{SUCCESS}, the generation halts. If \texttt{FAIL}, it generates a \texttt{suggestion} indicating the direction for parameter correction (e.g., \textit{``The target dictates 'backward translation', but the current velocity $[0, -1, 0]$ drives it downward. Velocity vector needs correction.''}).

\paragraph{Phase 2: Configuration Refinement with Hard-Rule Priors.}
To prevent the LLM from hallucinating invalid physical states (e.g., modifying global grid constraints or flipping coordinate axes), we enforce a strict \textbf{Hard-Rule Prioritization} protocol during the JSON rewrite.

The LLM is prompted with a deterministic constraint ledger, which mandates:
\begin{itemize}[leftmargin=*]
    \item \textbf{Directional Rigidity:} Base vectors are immutable. Composite directions must be synthesized via vector addition and strictly normalized ($||\mathbf{v}||_2 = 1$).
    \item \textbf{Skill-Specific Constraints:} If the action is classified as \texttt{scale} (deformation), the LLM must generate exactly two paired, symmetrically opposed \texttt{enforce\_particle\_translation} constraints. If classified as \texttt{impulse}, it triggers a \texttt{particle\_impulse} constraint bounded by an extremely brief time window.
    \item \textbf{Fallback Protocol:} If the LLM's intuition conflicts with the predefined directional mapping or feedback constraints, the hard-coded physical mappings take absolute precedence.
\end{itemize}

By restricting the LLM's degrees of freedom to only the valid parameter subspaces defined in our Force Field Skill Library, the Refine Agent guarantees both algorithmic convergence and numerical stability across multiple optimization iterations.

\subsection{Iteration Limit and Computational Efficiency}
\label{subsec:iteration_limit}

To rigorously balance the physical fidelity of the generated 4D dynamics with the overall computational overhead, we empirically establish a maximum limit of $3$ refinement iterations for the simulation-in-the-loop pipeline. 

Our extensive experiments in Table\ref{tab:iteration_clipsim} indicate that the Refine Agent typically achieves task convergence within this threshold. The initial generation (Iteration 0) inherently provides a strong geometric and physical baseline via the zero-shot Semantic Agent. Subsequent refinement steps (Iterations 1 and 2) efficiently rectify localized physical inconsistencies---such as misaligned directional vectors or disproportionate force magnitudes---based on the closed-loop visual feedback. 

We observe that permitting more than three iterations yields marginal improvements in the objective success rate, yet it linearly exacerbates the total generation time (which heavily comprises LLM inference, MPM forward simulation, and rendering). Therefore, capping the maximum iterations at $3$ acts as an optimal heuristic. It structurally guarantees a high success rate for complex interactive prompts while strictly minimizing the inference latency, rendering the framework highly pragmatic for real-world 4D synthesis tasks.

\begin{table}[h]
\centering
\caption{Quantitative impact of refinement iterations on the ClipSim metric. The performance gains plateau after the 3rd iteration, justifying our empirical iteration limit.}
\label{tab:iteration_clipsim}
\begin{tabular}{cc}
\toprule
\textbf{Refinement Iteration} & \textbf{ClipSim Score ($\uparrow$)} \\
\midrule
1 & 0.2347 \\
2 & 0.2678 \\
3 & \textbf{0.3160} \\
4 & 0.3159 \\
5 & 0.3161 \\
\bottomrule
\end{tabular}
\end{table}

\section{Fundamentals of Material Point Method}
\label{sec:mpm_fundamentals}

To ensure high-fidelity and physically grounded 4D synthesis, the forward execution engine of our framework is powered by the Material Point Method (MPM). MPM is a hybrid Lagrangian-Eulerian approach that effectively couples the advantages of particle-based tracking and grid-based computation, making it exceptionally robust for simulating large deformations, elastoplasticity, and complex topological changes (e.g., tearing or merging).

In our implementation, we leverage the highly parallelized \textit{NVIDIA Warp} framework to achieve high-performance GPU acceleration. The core MPM solver integrates the Affine Particle-in-Cell (APIC) transfer scheme to preserve angular momentum and reduce numerical dissipation.

The standard execution loop within a single time sub-step $\Delta t_{sub}$ consists of the following key phases:

\subsection{Particle-to-Grid (P2G) Transfer}
The continuum body is discretized into a set of Lagrangian particles, each carrying mass $m_p$, position $\mathbf{x}_p$, velocity $\mathbf{v}_p$, volume $V_p$, and the deformation gradient $\mathbf{F}_p$. In the P2G phase, the mass and momentum of particles are transferred to the background Eulerian grid nodes $i$ using tricubic interpolation shape functions $N_i(\mathbf{x}_p)$ and their gradients $\nabla N_i(\mathbf{x}_p)$:
\begin{equation}
m_i = \sum_p m_p N_i(\mathbf{x}_p)
\end{equation}
\begin{equation}
(m\mathbf{v})_i = \sum_p m_p \left( \mathbf{v}_p + \mathbf{C}_p (\mathbf{x}_i - \mathbf{x}_p) \right) N_i(\mathbf{x}_p)
\end{equation}
where $\mathbf{C}_p$ is the affine velocity matrix from the APIC formulation, which is crucial for reducing angular momentum damping.

\subsection{Grid Momentum Update and Force Application}
Once the mass and momentum are mapped to the grid, the nodal velocities are updated by solving Newton's second law. This step is where our \textit{Force Field Skill Library} (detailed in Appendix \ref{sec:force_field_skill}) takes effect. 

The velocity update equation is:
\begin{equation}
\mathbf{v}_i^{t+\Delta t} = \mathbf{v}_i^t + \Delta t m_i^{-1} \left( \mathbf{f}_i^{int} + \mathbf{f}_i^{ext}(\Theta_k) \right)
\end{equation}
Here, $\mathbf{f}_i^{ext}(\Theta_k)$ represents the external environmental forces dynamically configured by the LLM agents. The internal elastic force $\mathbf{f}_i^{int}$ is computed based on the Cauchy stress $\boldsymbol{\sigma}_p$:
\begin{equation}
\mathbf{f}_i^{int} = - \sum_p V_p \boldsymbol{\sigma}_p \nabla N_i(\mathbf{x}_p)
\end{equation}

\subsection{Constitutive Models and Return Mapping}
Our framework supports a versatile set of elastoplastic constitutive models to simulate diverse physical materials (e.g., jelly, metal, sand, and foam). The Cauchy stress $\boldsymbol{\sigma}_p$ is derived from the elastic part of the deformation gradient $\mathbf{F}_p^E$. 

Depending on the assigned material type, we apply specific hyperelastic energy density functions and their corresponding return mapping algorithms:
\begin{itemize}[leftmargin=*]
    \item \textbf{St. Venant-Kirchhoff (StVK) and Neo-Hookean:} Used for general hyperelastic materials, with Kirchhoff stress $\boldsymbol{\tau} = J \boldsymbol{\sigma}$ computed via SVD decomposition.
    \item \textbf{Von Mises with Damage Softening:} Employed for yielding materials like metal or plasticine. We implement a Von Mises return mapping scheme that projects the trial stress back to the yield surface. A softening coefficient is integrated to degrade the yield stress and Lamé parameters ($\mu, \lambda$) upon severe plastic deformation, enabling realistic material fracture.
    \item \textbf{Drucker-Prager:} Specifically implemented for granular materials like sand, where the yield criterion is dependent on the hydrostatic pressure (trace of the stress tensor).
\end{itemize}

\subsection{Grid-to-Particle (G2P) Transfer and Advection}
After the grid velocities $\mathbf{v}_i^{t+\Delta t}$ are updated and grid-based boundary conditions (e.g., sticky, slip, or frictional collisions) are enforced, the kinematic states are interpolated back to the particles:
\begin{equation}
\mathbf{v}_p^{t+\Delta t} = \sum_i \mathbf{v}_i^{t+\Delta t} N_i(\mathbf{x}_p)
\end{equation}
\begin{equation}
\mathbf{x}_p^{t+\Delta t} = \mathbf{x}_p^t + \Delta t \mathbf{v}_p^{t+\Delta t}
\end{equation}
Finally, the trial deformation gradient is advected as $\mathbf{F}_p^{trial} = (\mathbf{I} + \Delta t \nabla \mathbf{v}_p) \mathbf{F}_p^t$, which is subsequently fed into the return mapping algorithms in the next sub-step to update the elastic deformation $\mathbf{F}_p^E$.

\subsection{Base Simulation Hyperparameters}
\label{subsec:mpm_hyperparameters}

While the extrinsic dynamics are dynamically injected by the LLM (as detailed in Appendix \ref{sec:force_field_skill}), the underlying MPM simulator relies on a robust set of predefined base parameters to guarantee numerical stability and physical fidelity throughout the execution loop. In our framework, these hyper-parameters are configured as follows:

\begin{itemize}[leftmargin=*]
    \item \textbf{Temporal and Spatial Resolution:} To prevent numerical explosion (e.g., violating the Courant-Friedrichs-Lewy (CFL) condition) during extreme deformations, the simulation operates at a highly granular spatial resolution (\texttt{n\_grid} $= 100$) and an extremely fine temporal sub-step (\texttt{substep\_dt} $= 2\times 10^{-4}$ s). The states are then accumulated and rendered at a macroscopic frame interval (\texttt{frame\_dt} $= 4\times 10^{-2}$ s).
    \item \textbf{Numerical Damping and Stability:} To simulate realistic energy dissipation and prevent infinite oscillation, we introduce explicit damping factors. \texttt{grid\_v\_damping\_scale} (e.g., $0.9995$) slightly decays the grid velocity globally. Additionally, \texttt{rpic\_damping} ($0.5$) blends the standard APIC and PIC transfers, trading off a fraction of angular momentum preservation for enhanced numerical stability during chaotic collisions.
    \item \textbf{Base Boundary Conditions:} Predefined spatial boundaries (e.g., \texttt{bounding\_box}) and static kinematic constraints are enforced before the LLM's dynamic force fields are applied, ensuring particles remain within the valid simulation domain.
\end{itemize}

\section{Evaluation Metrics and User Study}
\label{sec:eval_and_user_study}

To comprehensively evaluate the quality of the synthesized 4D physical dynamics, we employ both objective quantitative metrics and rigorous subjective human evaluations.

\subsection{Mathematical Formulation of CLIPsim}
\label{subsec:clipsim}
The CLIPsim metric is utilized to objectively measure the semantic alignment between the user's text prompt and the generated video sequence. Let the generated video be denoted as $\mathbf{V} = \{v_1, v_2, \dots, v_N\}$, comprising $N$ frames, and the semantic text prompt be $\mathcal{P}$. We extract the visual feature embeddings $E_I(v_i) \in \mathbb{R}^d$ and the text feature embedding $E_T(\mathcal{P}) \in \mathbb{R}^d$ using a pre-trained CLIP vision-language model. 

The frame-level CLIPsim score is computed as the cosine similarity between the text and image embeddings. To assess the temporal-global alignment, the final $CLIP_{sim}$ is defined as the temporal average of these similarities across the entire video:
\begin{equation}
    CLIP_{sim}(\mathbf{V}, \mathcal{P}) = \frac{1}{N} \sum_{i=1}^{N} \frac{E_I(v_i) \cdot E_T(\mathcal{P})}{||E_I(v_i)||_2 ||E_T(\mathcal{P})||_2}
\end{equation}
A higher $CLIP_{sim}$ indicates that the dynamic trajectory and object states within the video accurately reflect the semantic intent (e.g., ``stretching'', ``squeezing'', or ``colliding'') specified in the prompt.

\subsection{Subjective Human Evaluation (User Study)}
\label{subsec:user_study}
Given the highly complex nature of 4D physical synthesis, objective metrics alone cannot fully capture the nuances of physical realism and material deformation. Therefore, we conducted a rigorous user study using a 3-Alternative Forced-Choice (3AFC) methodology.

\paragraph{Participants and Setup.}
We invited $20$ expert researchers specialized in Computer Vision and Computer Graphics. Their domain expertise ensures a highly critical and professional assessment of physical fidelity, rendering artifacts, and dynamic consistency, mitigating the subjective bias often found in crowdsourced evaluations.

\paragraph{Methodology.}
The participants were presented with 9 sets of generated videos. Each set contained three unlabeled, randomly shuffled videos generated from the same initial state and prompt, using three different settings in Sec\ref{subsec:compare}.

For each set, participants were asked to select the absolute best video across three distinct criteria:
\begin{itemize}[leftmargin=*]
    \item \textbf{Text-to-Action Alignment:} Which video executes the physical action most accurately according to the user's textual prompt?
    \item \textbf{Physical Fidelity:} Which video demonstrates the most realistic material deformation and rigid/soft body dynamics without obvious artifacts?
    \item \textbf{Overall Preference:} Considering both visual quality and physical accuracy, which video is the most preferred?
\end{itemize}

\section{Limitations and Future Work}
\label{sec:limitations}

While PhysAgent demonstrates strong capabilities in translating semantic intents into 4D physical dynamics, our current framework naturally presents several exciting avenues for future research and refinement:

\paragraph{Granularity of Force Field Coverage.}
Fundamentally, our MPM simulator applies external vector fields guided by semantic object masks (provided by SAM 2) and bounding geometries. This object-level force coverage is highly effective for global interactions, but presents opportunities for further refinement when handling extremely fine-grained, localized dynamics (e.g., precisely pulling a single thin string on a complex woven fabric). Currently, the mask-based field tends to distribute energy broadly. Future iterations could integrate part-level segmentation models or hierarchical semantic mapping to enable precise, localized force primitives, paving the way for intricate scenarios like fluid-solid coupling.

\paragraph{Efficiency-Convergence Trade-offs in Multi-Stage Scenarios.}
To ensure pragmatic computational efficiency and maintain an interactive generation pace, we empirically cap the Refine Agent at 3 iterations (as discussed in Appendix \ref{subsec:iteration_limit}). While this heuristic guarantees high success rates for standard prompts, highly non-linear or multi-stage physical intents (e.g., \textit{``the object is severely compressed, violently twisted, and then rebounds''}) yield a highly non-convex parameter space. In such complex cases, halting at 3 iterations may sometimes terminate the search before reaching the global optimal boundary conditions. Future frameworks could explore adaptive iteration limits based on real-time convergence metrics, or introduce LLM-driven prompt decomposition to break complex sequences into manageable, sequential sub-tasks.

\paragraph{Vulnerability to Extreme Occlusion and Textureless Surfaces.} 
Our closed-loop visual perception relies heavily on point-tracking algorithms (CoTracker3) and monocular depth estimation. When simulating extreme material yielding (e.g., a jelly completely crushed into flat pieces) or handling textureless surfaces (e.g., a perfectly smooth mirror ball), point trackers inevitably lose trajectories due to severe self-occlusion or lack of visual features. This degrades the quality of the Semantic Feedback, subsequently impairing the LLM's refinement judgments.